\documentclass[12pt, A4]{article}

\usepackage[utf8]{inputenc}
\usepackage[left=2cm, right=2cm, top=2cm, bottom=2cm]{geometry}
\usepackage{titling}
\usepackage{graphicx}

\usepackage{parskip} 

\newcommand{\subtitle}[1]{%
  \posttitle{%
    \par\end{center}
    \begin{center}\Large#1\end{center}
    \vskip0.5em}%
}

\usepackage{setspace}

\usepackage[sort&compress]{natbib}
\usepackage{rotating}
\usepackage{bbm}
\usepackage{latexsym}

\usepackage{algorithm}
\usepackage{algpseudocode}

\usepackage{xcolor}

\usepackage{amsmath,amssymb,mathtools}
\usepackage{bbm}
\usepackage{xfrac}

\usepackage{amsthm}

\usepackage[colorlinks]{hyperref}

\definecolor{mygrey}{RGB}{125, 125, 125}
\definecolor{mygrey2}{RGB}{100, 100, 100}
\definecolor{mygreen}{RGB}{0, 120, 0}
\definecolor{myred}{RGB}{120, 0, 0}
\definecolor{myblue}{RGB}{0, 0, 120}

\hypersetup{
    colorlinks,
    linkcolor={red!50!black},
    citecolor={mygrey2},
    urlcolor={blue!80!black}
}

\usepackage{pifont}
\usepackage{setspace}

\newcommand{\appropto}{\mathrel{\vcenter{
  \offinterlineskip\halign{\hfil$##$\cr
    \propto\cr\noalign{\kern2pt}\sim\cr\noalign{\kern-2pt}}}}}

\usepackage{pgfplotstable}
\usepackage{booktabs}
\pgfplotsset{compat=1.18}

\title{Meta-learning as a principle for human-like visual representations\vspace{0.2cm}}

\author{Can Demircan$^{1,*}$, Marcel Binz${^1}$, Alireza Modirshanechi$^{1,2}$, and Eric Schulz$^{1}$\vspace{0.2cm}}
\date{\vspace{-1cm}}

\begin{document}

\maketitle
\thispagestyle{empty}
\begin{center}
\small{$^1$ Helmholtz Munich, Munich, Germany}\\
\small{$^2$ Max Planck Institute for Biological Cybernetics, Tübingen, Germany}\\
\small{$^*$ Corresponding author: can.demircan@helmholtz-munich.de}\\
\end{center}
\vspace{0.5cm}

\begin{abstract}
The structure of human visual representations underpins our capacity for adaptive behaviour. While pretrained neural networks model human visual representations with unprecedented success, a large discrepancy remains. We propose one reason: these networks optimise a single fixed objective, whereas human representations must support open-ended tasks. We hypothesise this flexibility arises from meta-learning (learning to learn), a pressure shaping representations to acquire new tasks from few observations. To test this, we train a sequence model, without any supervision from human data, across thousands of semantically rich tasks mapping images to high-level concepts. Compared to their pretrained base encoders, meta-learned representations better predict human similarity judgements, semantic rule learning, and high-level visual cortex. Behavioural gains depend on disentangled, high-level task distributions, while brain alignment is driven primarily by the learning-to-learn pressure. Our results suggest the flexibility of human visual representations reflects the functional demand to learn new semantic relationships on the fly.
\end{abstract}

\newpage
\section*{Introduction}

Understanding how the mind represents the visual world is a central goal of cognitive science, as these representations underpin our capacity for flexible behaviour. Over the last decade, off-the-shelf neural networks have become powerful tools for pursuing this goal: their internal visual representations, how they encode and organise visual information, increasingly align with behavioural and brain data, with distances between model representations predicting human similarity judgments and responses in visual cortex~\cite{kriegeskorte_deep_2015, sucholutsky_getting_2025, yamins_performance-optimized_2014, schrimpf_brain-score_2018, sucholutsky_alignment_2023}. This progress has clarified what makes a good model in practice: large-scale architectures and diverse training datasets typically yield better alignment with humans~\cite{muttenthaler_human_2022,conwell_large-scale_2024, demircan_evaluating_2024}, and targeted fine-tuning directly on human data can bring models closer still~\cite{muttenthaler_aligning_2025,fel_harmonizing_2022,fu_dreamsim_2023,sundaram_when_2024, roads_enriching_2021, muttenthaler_improving_2023}. These are remarkable accomplishments, but they largely leave open how and why human representations emerge: alignment can be engineered by training on human data, yet this does not explain why human representations, which arise without such supervision, are naturally organised to support tasks ranging from categorisation to rapid rule learning.

To address this, we hypothesise that the structural alignment between human and machine visual representations is not simply a by-product of scale or standard feature learning, but a consequence of the functional demand to be repurposed for novel requirements. We investigate this through the lens of meta-learning (learning to learn), which provides a normative framework for how these inductive biases emerge~\cite{binz_meta-learned_2024, griffiths_doing_2019}. In meta-learning, an agent is repeatedly confronted with new tasks drawn from a broader family, yielding strategies and representations that are general, flexible, and reusable. By shifting the focus away from explicitly fine-tuning models on static human data, meta-learning instead requires a model to rapidly adapt to unseen rules from limited observations. This framework has been used to account for a range of cognitive phenomena, from prefrontal learning dynamics~\cite{wang_prefrontal_2018} to human-like compositional generalisation in language~\cite{lake_human-like_2023}. More recently, meta-learning over ecologically grounded task distributions has been shown to capture human-like behaviour in decision-making by internalising the statistical regularities of naturalistic environments~\cite{jagadish_human-like_2024, jagadish2025meta}. However, this line of work has focused on modelling decisions and cognitive strategies rather than the representations on which they operate; whether the same learning-to-learn pressure shapes human visual representations themselves remains an open question.

Investigating whether meta-learning can capture these demands requires a task-agnostic training environment rich enough to reflect the diversity of human conceptual space. Human vision is exceptionally versatile: depending on the context, an apple can be framed by its visual properties (colour) for a painting, its abstract category (fruit) when hungry, or its arbitrary role in an unfamiliar task. Because we cannot predict which of these dimensions will be relevant at any given moment, our internal representations must remain flexible. To simulate this pressure, we must generate thousands of semantically meaningful tasks at scale. To generate such tasks at scale, we leverage Sparse Autoencoders (SAEs), which decompose the entangled internal activations of pretrained networks into a much larger set of sparse, interpretable units~\cite{cunningham_sparse_2023, bricken_towards_2023, elhage_toy_2022, bau_network_2017}. Unlike individual activations, these latents align with high-level human concepts, such as fabric items, small animals, and kitchen-related objects, providing a structured and diverse conceptual vocabulary from which to build tasks.

\begin{figure}[h!]
    \centering 
    \includegraphics[width=\textwidth]{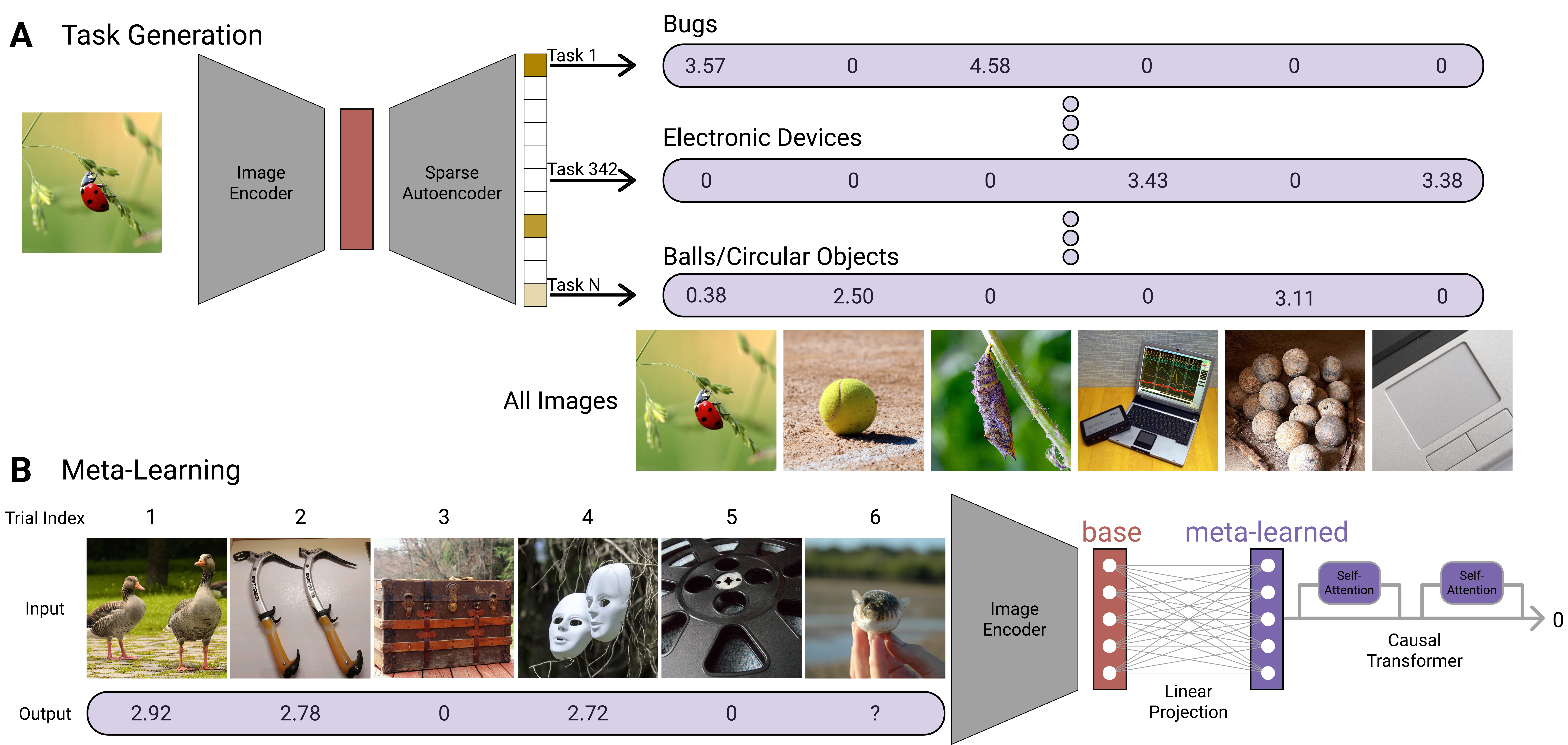}
    \vspace{-1.2cm}
    \singlespacing  \caption{ \footnotesize
\textbf{Overview of the meta-learning framework.} \textbf{A.} Semantically rich tasks are generated at scale using Sparse Autoencoders (SAEs) applied to a pretrained CLIP model. Each SAE latent defines a task over images: images with non-zero activations for a given latent are positive examples, with activation magnitude serving as the graded target. Example tasks are shown with images from the THINGSplus database~\cite{stoinski_thingsplus_2024}, displaying real activation values to illustrate that individual latents capture coherent high-level concepts. Task index numbers shown are arbitrary. \textbf{B.} A causal Transformer is trained to predict semantically rich tasks from sequences of images. At each trial, the model receives an image representation from a frozen pretrained encoder together with the previous trial's outcome, and produces two predictions: whether the current image is a positive instance of the sampled task, and the task's activation value. For downstream evaluations, we extract static image representations by applying the learned linear projection to encoder features independently per image, without sequential context (labelled "meta-learned" in the schematic, contrasted with the frozen encoder features labelled "base").}
    \label{fig:overview}
\end{figure}

Using this vocabulary, we formulated thousands of few-shot learning tasks, where each task is defined by a specific SAE latent (\autoref{fig:overview}A). We then train a causal Transformer to meta-learn these tasks: at each trial, the model observes an image and must predict the task's output for that image (\autoref{fig:overview}B). The inputs are features from several state-of-the-art frozen image encoders (SigLIP2~\cite{tschannen_siglip_2025}, Masked Autoencoder (MAE)~\cite{fan_scaling_2025, he_masked_2022}, DINOv3~\cite{simeoni_dinov3_2025}, and CLIP~\cite{ilharco_openclip_2021,radford_learning_2021}). Meta-learning reshapes these features only through a learned linear projection applied to each image. We take this projected, context-independent representation as the meta-learned representation and compare it against the original frozen (base) features. Crucially, this training process is entirely unsupervised with respect to human data. To verify that any alignment gains come from the learning-to-learn pressure specifically rather than mere exposure to semantically rich tasks, we also compare against a multitask model trained on the same task distribution without episodic structure.

We evaluated these models across five publicly available behavioural and neuroimaging datasets. Meta-learned representations were better predictors of human similarity judgements than those derived from base models and better captured how humans learn new semantic rules from limited observations. They also showed improved alignment with the human visual cortex, particularly in high-level category-selective regions. These effects depended on the functional requirement to learn over tasks defined by disentangled, high-level concepts.

\section*{Results}
\subsection*{Meta-learned representations are better models of human behaviour}

\begin{figure}[t!]
    \centering 
    \includegraphics[width=\textwidth]{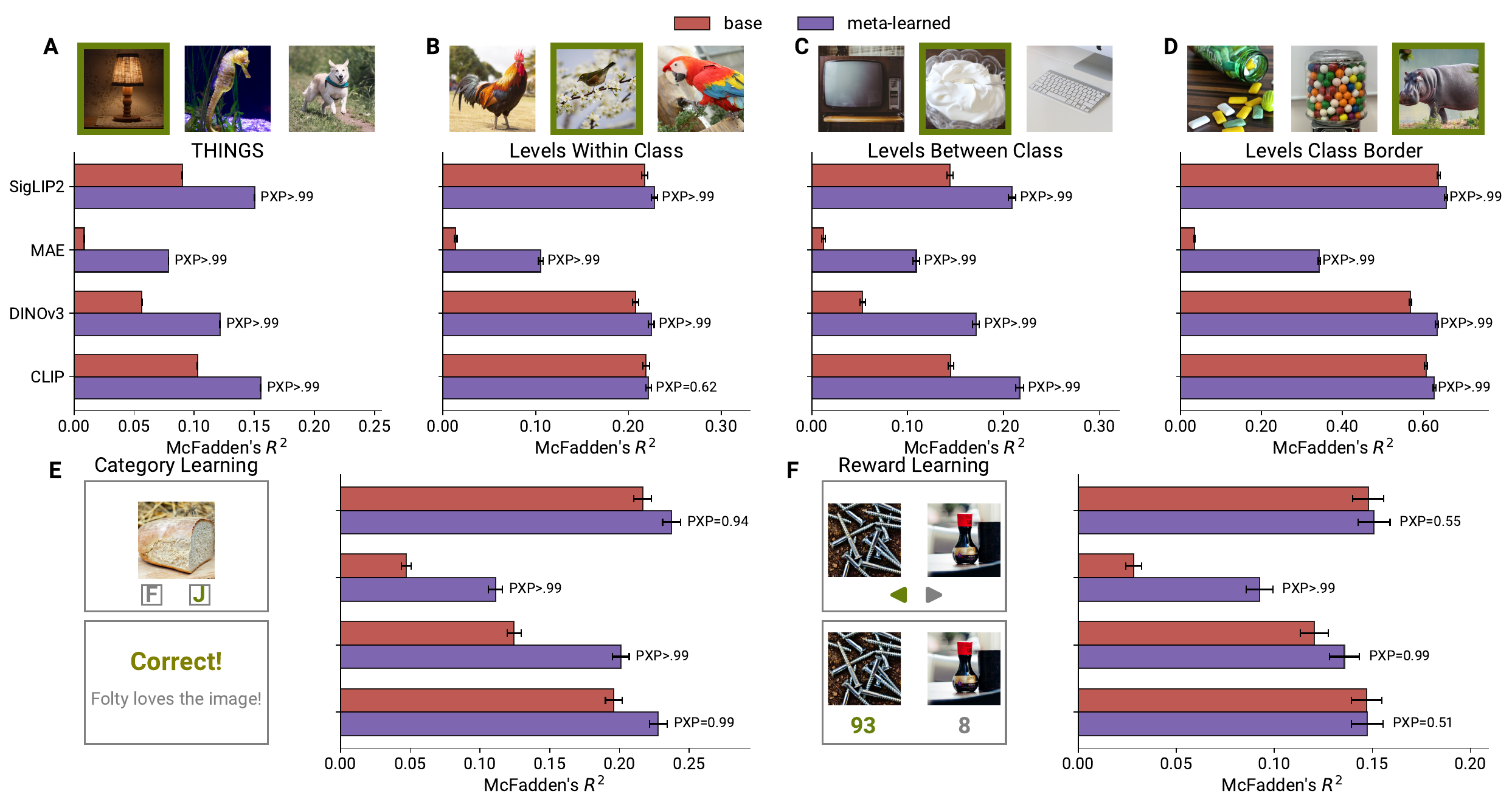}
    \vspace{-1.2cm}
    \singlespacing  \caption{ \footnotesize
    \textbf{Meta-learning over semantically rich tasks aligns visual representations with human behaviour.} \textbf{A}. Behavioural alignment with human odd-one-out choices on the THINGS dataset, quantified by McFadden's $R^2$ based on the negative log-likelihood (NLL) of human choices ($R^2 = 1$: perfect prediction; $R^2 = 0$: chance-level). \textbf{B–D}. Behavioural alignment on the Levels dataset across different hierarchical depths, including within-class (\textbf{B}), between-class (\textbf{C}), and class-border (\textbf{D}) trials. \textbf{E, F}. Behavioural alignment on sequential learning tasks. \textbf{E}. Results for the category-learning task, where participants learned hidden semantic rules. \textbf{F}. Results for the reward-learning task, where rewards were tied to latent semantic dimensions. Text annotations indicate the protected exceedance probability (PXP), representing the probability that the meta-learned model is a better fit for the human data than the base model. Error bars represent the standard error of the mean (SEM).
    }
    \label{fig:evals}
\end{figure}

We train a causal Transformer to learn novel semantically rich tasks in context from short sequences of images. At its input, the Transformer learns a linear projection that maps each encoder feature into a space suited for in-context learning. After training, we obtain a meta-learned representation for any single image by applying this learned projection to its encoder features independently, without any sequence context (the meta-learned block in \autoref{fig:overview}). We contrast this against the base representation, the unmodified encoder features. The meta-learned representation is therefore a static, linear transformation of the base features, which lets us ask whether the format induced by learning-to-learn is itself more human-aligned, independent of any in-context adaptation at test time. We apply this pipeline to four frozen encoders spanning contrastive (SigLIP2, CLIP), self-supervised distillation (DINOv3), and masked-autoencoding (MAE) objectives, and evaluate the resulting representations across five publicly available behavioural and neuroimaging datasets. In almost all cases, the meta-learned representations were better models of human behaviour (\autoref{fig:evals}).

\paragraph{Similarity judgements} We first evaluated the representations on the THINGS odd-one-out similarity judgement dataset~\cite{hebart_revealing_2020,hebart_things-data_2023}, a large-scale collection of over $4.6$ million triplet choices in which participants were asked to identify which of three images was the odd one out. This dataset provides a good test-bed for capturing humans' semantic intuitions about the visual world.

To turn a representation into a behavioural prediction, we used the following procedure: For each triplet, we computed the pairwise cosine similarity between the three image representations; the predicted odd-one-out is the image whose removal leaves the most similar remaining pair. These similarities are converted into per-trial choice probabilities via a softmax with a per-participant temperature fit by cross-validation. Behavioural alignment is quantified using McFadden's $R^2$ based on the negative log-likelihood (NLL) of the human choices, which normalises the model's likelihood against chance where $R^2 = 1$ corresponds to perfect prediction of human choices, and $R^2 = 0$ to chance-level. Across all tested encoders (SigLIP2, MAE, DINOv3, and CLIP), the meta-learned representations were better models of human judgements than the base models (PXP $> .99$, \autoref{fig:evals}A).

We then tested the representations on another odd-one-out dataset named Levels~\cite{muttenthaler_aligning_2025,muttenthaler_levels_2024}, where different triplets correspond to the assessment of different levels of the semantic hierarchy. On one extreme are within-class trials, where all the images come from the same category. On the other end are between-class trials, where each image comes from a different category; in between are class-border trials, where two images come from the same category and the third from a different one. Alignment was quantified using the same cosine-similarity procedure and McFadden's $R^2$ as above. Across all levels of abstraction and all tested encoders, we found that meta-learned representations were better models of human behaviour (PXP $> .99$) in all but one comparison, where the CLIP encoder did not show a clear difference for the within-class comparison (PXP $= 0.62$, \autoref{fig:evals}B-D).

\paragraph{Learning tasks} Human visual representations are not only used for making static judgements but to quickly learn new functional relationships after only a few observations. We compared the meta-learned representations against the base representations in naturalistic category- and reward-learning tasks~\cite{demircan_evaluating_2024} to test how well they capture human visual learning.

To evaluate a given set of representations, we simulated a participant by fitting a simple online learner on top of those representations: at each trial $t$, the learner is fit to the participant's feedback on the preceding trials $1{:}t{-}1$ and is used to predict the choice on trial $t$. The richer the representations, the easier it is for a minimal learner on top of them to match human trial-by-trial choices. Behavioural alignment is again quantified using McFadden's $R^2$ on the NLL of human choices.

In the category-learning task, participants were presented with one image at a time and asked to predict which of two categories the image belonged to, receiving feedback on each trial. Unbeknown to them, category membership was assigned based on a hidden rule (e.g., whether the object is kitchen-related or not). We simulated this process with an online L2-regularised logistic regression over the representations. Across all the tested encoders, meta-learned representations were more likely to produce human choices than base model representations (PXP $= .94$ for SigLIP2, PXP $> .99$ for MAE, DINOv3, and CLIP; see \autoref{fig:evals}E). 

In the reward-learning task, participants saw two images at a time and were instructed to select the one they believed was more rewarding, receiving feedback on each trial. As in the category-learning task, rewards were assigned based on a semantically meaningful rule (e.g., whether one option is more metallic than the other). We simulated this with an online Bayesian ridge regression over the representations, using the difference in predicted reward as the choice logit. For MAE and DINOv3, meta-learning yielded more human-like representations (PXP $\geq .99$), whereas we observed no meaningful difference for SigLIP2 and CLIP (PXP $= .55$ and PXP $= .51$ respectively; see \autoref{fig:evals}F).

Taken together, these results demonstrate that meta-learning on a diverse distribution of semantically rich tasks reorganises visual features into a format more accessible for human-like judgements and learning.

\subsection*{Meta-learning, disentanglement, and abstraction drive human-like representations}
 
\begin{figure}[t!]
    \centering 
    \includegraphics[width=\textwidth]{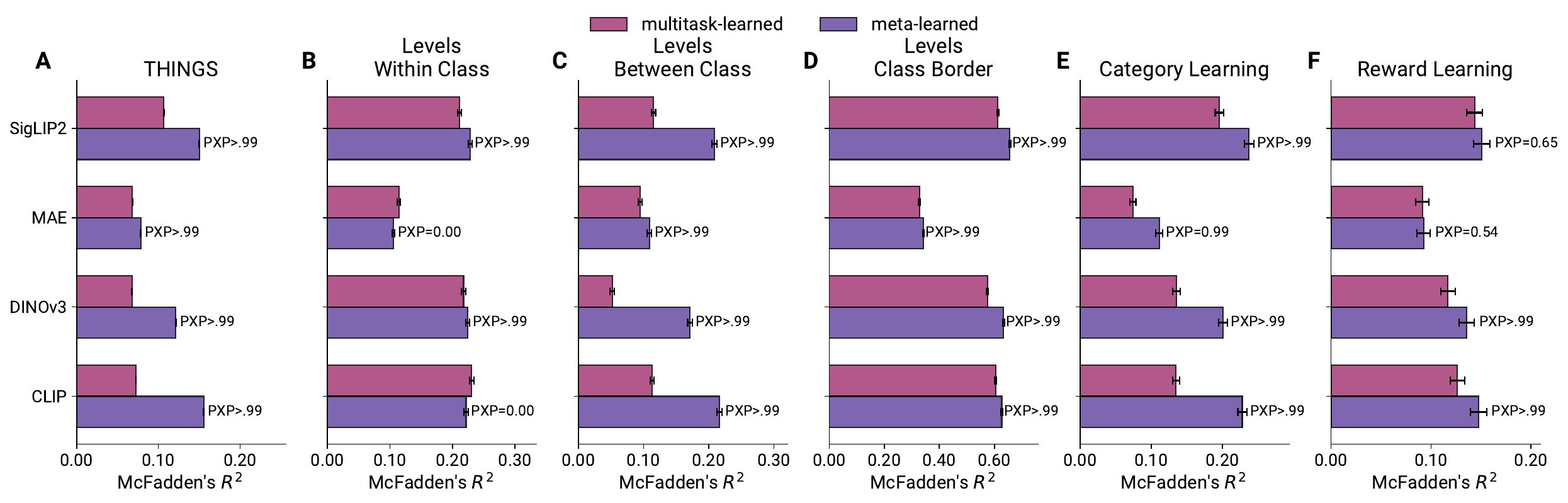}
    \vspace{-1.2cm}
    \singlespacing  \caption{ \footnotesize
    \textbf{Meta-learning outperforms multitask learning on the same task distribution.} Comparison of meta-learned representations (purple) against a non-sequential multitask model (pink) trained on the same task distribution but without episodic structure, processing each image independently via per-task linear heads. The multitask model receives the same image-task supervision as the meta-learner, isolating the contribution of the learning-to-learn pressure itself. Behavioural alignment is shown on the THINGS odd-one-out task (\textbf{A}), Levels Within Class (\textbf{B}), Between Class (\textbf{C}), Class Border (\textbf{D}), category learning (\textbf{E}), and reward learning (\textbf{F}), measured by McFadden's $R^2$. Text labels indicate the protected exceedance probability (PXP). Purple bars are the same meta-learned results shown in \autoref{fig:evals}. Error bars represent the SEM.
    }
    \label{fig:meta}
\end{figure}

\begin{figure}[t!]
    \centering 
    \includegraphics[width=\textwidth]{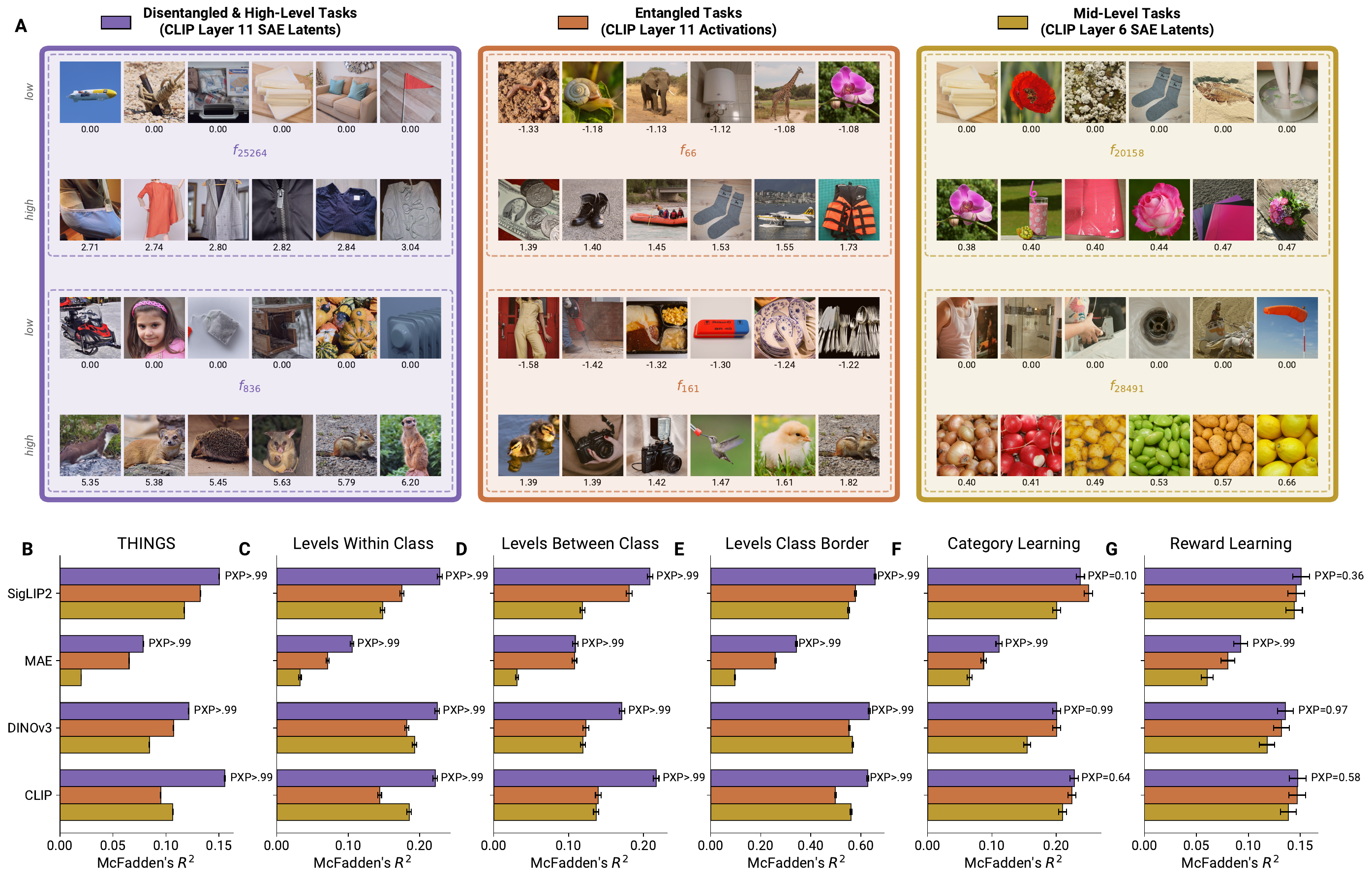}
    \vspace{-1.2cm}
    \singlespacing  \caption{ \footnotesize
    \textbf{Disentangled and high-level tasks drive human alignment.} \textbf{A}. Visualisations of the three task distributions used for meta-learning: high-level disentangled tasks (SAE latents from CLIP Layer 11; left), high-level entangled tasks (raw residual stream from CLIP Layer 11; middle), and mid-level disentangled tasks (SAE latents from CLIP Layer 6; right). Examples show images that maximally and minimally activate example tasks within each distribution. \textbf{B–G}. Comparative behavioural alignment of models trained on these distributions across human behavioural tasks, including the THINGS odd-one-out task (\textbf{B}), Levels dataset (\textbf{C–E}), category learning (\textbf{F}), and reward learning (\textbf{G}), quantified by McFadden's $R^2$. Purple bars are the same meta-learned results reported in \autoref{fig:evals}; orange and yellow bars show the two alternative task distributions. Across most tasks and backbones, meta-learning on high-level disentangled tasks (purple) significantly outperforms training on entangled (orange) or lower-level (yellow) tasks. Text labels indicate the protected exceedance probability (PXP). Error bars represent the SEM.
    }
    \label{fig:distributions}
\end{figure}

Having established that meta-learning produces systematic gains over the base models, we now dissect which properties of the setup are responsible. We address this question from three complementary angles: meta-learning versus multitask learning, the role of the disentangled nature of the tasks, and the role of their abstraction level. In each case, we swap out one ingredient at a time and measure how alignment changes. In \autoref{fig:meta} and \autoref{fig:distributions}, the meta-learned bars correspond directly to the main results reported in \autoref{fig:evals}; the comparison bars show what happens when the corresponding ingredient is removed or changed.

\paragraph{Meta-learning versus multitask learning} A natural alternative to meta-learning is multitask learning~\cite{caruana_multitask_1997}: a model that is exposed to the same semantically rich tasks, but learns a fixed mapping from images to task outputs rather than being required to infer the task on the fly. If the gains we observed simply reflect exposure to a rich distribution of semantic tasks, then a multitask model trained on the same distribution should match the meta-learner in predicting human behaviour. If, on the other hand, the learning-to-learn pressure is doing the work, the meta-learner should win. To adjudicate, we trained a non-sequential multitask model on the same task distribution, with per-task linear heads that processed each image independently, without any sequence context or trial-by-trial feedback. This model receives exactly the same training signal as the meta-learner, but is never required to learn a task within an episode.

Across most behavioural comparisons, meta-learned representations significantly outperformed the multitask model (\autoref{fig:meta}). 
Multitask learning was competitive in only a small number of settings (Levels Within Class for MAE and CLIP, Reward Learning for MAE and SigLIP2). Outside these few cases, meta-learning consistently outperformed the multitask model, supporting the interpretation that the gains arise from the learning-to-learn pressure rather than task exposure alone.

\paragraph{The role of disentanglement} We next asked which properties of the task distribution matter. In the main analyses, we used high-level semantically rich tasks derived from SAE latents at layer 11 of the CLIP residual stream; these latents typically represent highly abstract and semantically interpretable concepts, such as fabric items or small animals (\autoref{fig:distributions}A left). To test the role of the SAE's sparse, interpretable structure, we trained a model on the raw residual stream representations of the OpenCLIP model at the same layer. This kept the level of visual abstraction constant while removing the disentangled structure of the SAE (\autoref{fig:distributions}A middle). Across almost all behavioural alignment evaluations, we found that models trained on sparse, disentangled SAE latents yielded superior alignment compared to those trained on entangled activations (PXP $\geq .99$ for all except SigLIP2 and CLIP comparisons in the learning tasks; see \autoref{fig:distributions}B-G). This performance gap suggests that meta-learning specifically over disentangled tasks is a critical driver for developing representations that match human semantic intuitions.

\paragraph{The role of abstraction} We further investigated whether the level of visual abstraction influenced the results by training a separate SAE on the residual stream of layer 6 of the same model. This allowed us to maintain the sparse structure of the tasks while shifting the distribution toward mid-level visual features, such as textures, repetitive patterns, and object parts (\autoref{fig:distributions}A right). In our comparisons, tasks derived from high-level features (Layer 11) were significantly more effective at inducing human-like representations than those from mid-level features (Layer 6) across nearly all evaluations (\autoref{fig:distributions}B-G).

Together, these three ablations show that the emergence of human-like visual representations through meta-learning is not a generic property of the training objective alone. It requires meta-learning (rather than multitask learning) and a task distribution that is sparse \emph{and} high-level. Removing any one of these ingredients weakens the gains substantially.

 \subsection*{Meta-learning improves alignment with the human brain}
 
\begin{figure}[t!]
    \centering 
    \includegraphics[width=\textwidth]{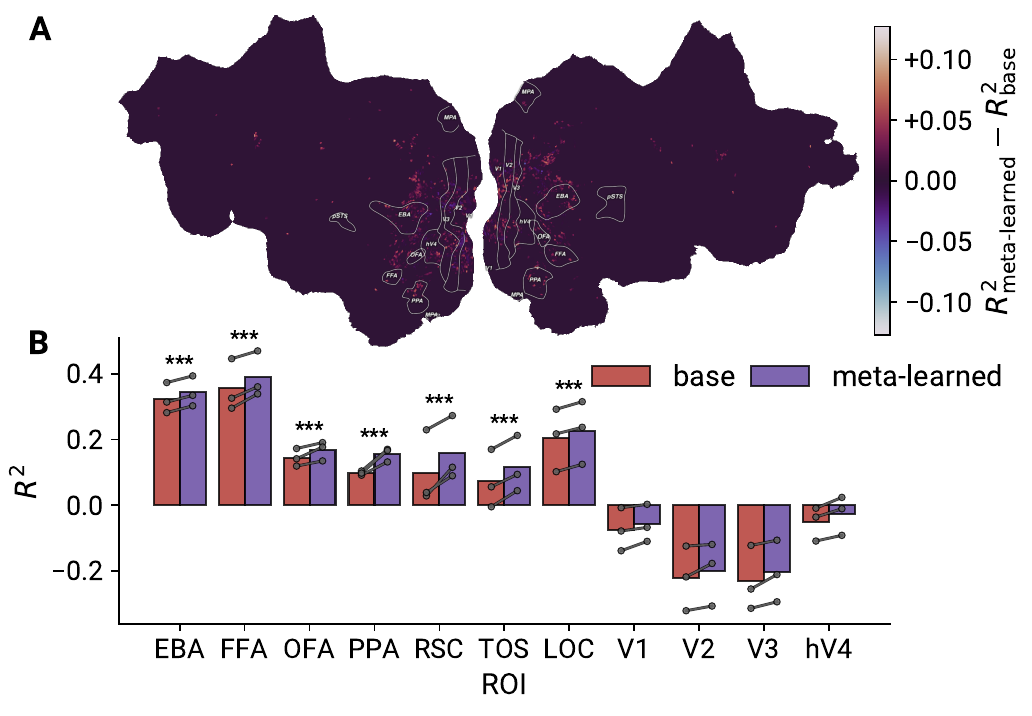}
    \vspace{-1.2cm}
    \singlespacing  \caption{ \footnotesize
    \textbf{Meta-learning improves alignment with the human visual cortex.} \textbf{A}. Whole-brain visualisation for Participant 1 showing the difference in noise-ceiling corrected predictive power ($R^2_{\text{meta-learned}} - R^2_{\text{base}}$) using the SigLIP2 backbone. Brighter colours indicate regions where meta-learning leads to a better fit of the fMRI data. \textbf{B}. Group-level predictive power ($R^2$) for base (red) and meta-learned (purple) SigLIP2 models across 11 regions of interest (ROIs) ($N=3$). Statistical significance is indicated by asterisks ($***$), determined by combining $p$-values from exhaustive sign-flip tests across participants using Fisher's method, with results corrected for multiple comparisons across ROIs using the Benjamini--Hochberg procedure. Results for other backbones show similar improvements and are provided in the Supplementary Fig.~S4. Meta-learning significantly improves neural alignment in category-selective areas while maintaining performance in early visual areas. Individual data points represent the average $R^2$ per participant.
    }
    \label{fig:brains}
\end{figure}

If this reorganisation is genuinely human-like, and not just a reshuffling that happens to fit behavioural benchmarks, it should also make the representations a better match for the neural populations that underlie human semantic vision. To test this, we evaluated our models using the THINGS-fMRI dataset ~\cite{hebart_things-data_2023}, which contains brain responses from three participants over $12$ sessions each, totalling $8{,}740$ unique images. We used ridge regression to predict voxel-wise activity across 11 regions of interest (ROIs) that span the human visual hierarchy, including early visual areas and high-level category-, object-, and scene-selective regions. In general, meta-learned representations showed a systematic increase in predictivity compared to their corresponding base models (\autoref{fig:brains}).

The improvements were broad and consistent across all tested high-level regions. We found significant increases in areas specialised for perceiving bodies and faces (EBA, FFA, OFA; $p < .001$), as well as in object- and scene-selective regions (LOC, PPA, RSC, TOS; $p < .001$). Early visual areas (V1, V2, V3, hV4) had a below-chance predictive accuracy both for base and meta-learned representations, so we did not conduct statistical tests for these ROIs. This is expected given that the high-level representations used as inputs likely lack the fine-grained visual information that early voxels are tuned to in a linearly decodable format.

The ablations of the previous section also replicate in the brain, though with differing strengths. Consistent with the behavioural picture, meta-learning yielded significantly higher predictive power than multitask learning across all high-level ROIs and all four encoders ($p < .001$; Supplementary Fig.~S3). The task-distribution ablations show a more heterogeneous picture: the high-level disentangled distribution was not systematically better than the entangled or mid-level alternatives across ROIs and encoders (Supplementary Fig.~S5). Together, this suggests that while both learning-to-learn and the format of the task distribution shape behaviour, brain alignment is driven primarily by the meta-learning pressure itself.

\subsection*{Meta-learning reorganises geometry in a human-like fashion}

\begin{figure}[t!]
    \centering 
    \includegraphics[width=\textwidth]{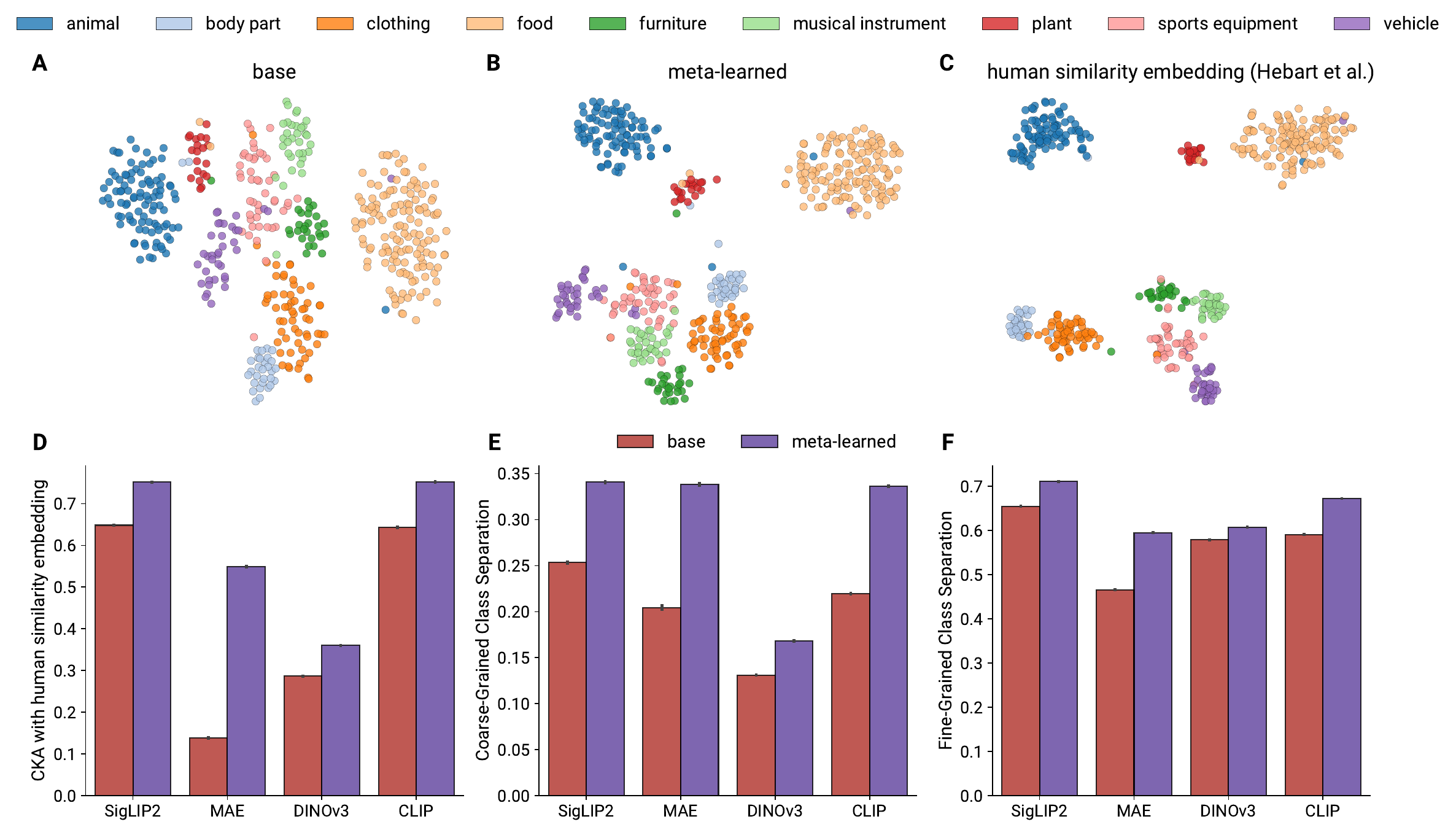}
    \vspace{-1.2cm}
    \singlespacing  \caption{ \footnotesize
    \textbf{Meta-learning reorganises representational geometry to mirror human conceptual structure.} \textbf{A–C}. t-SNE visualisations of THINGS images for (\textbf{A}) the SigLIP2 base model, (\textbf{B}) the meta-learned model, and (\textbf{C}) the 66-dimensional SPoSE human similarity embedding derived from the THINGS odd-one-out dataset. Meta-learning exaggerates the global separation between organic (e.g., animals, plants) and inorganic categories (e.g., furniture, vehicles), matching the structure of the human embedding. \textbf{D}. Alignment with the human similarity embedding via CKA. Meta-learned models (purple) show significantly higher alignment across all backbones compared to base models (red). \textbf{E, F}. Class separation for coarse-grained semantic categories (\textbf{E}) and fine-grained object categories (\textbf{F}). Meta-learning improves categorical distinctness at both levels of the semantic hierarchy. Error bars indicate standard errors from bootstrap resampling.}
    \label{fig:geoms}
\end{figure}

Lastly, we asked how meta-learning transforms the underlying structure of visual features. Although the meta-learned representation is only a linear projection of the base features, this projection is not distance-preserving. By reweighting and recombining feature dimensions, it changes which images sit near one another, and hence the relational structure that downstream cosine-similarity read-outs depend on. To make this restructuring visible, we projected the THINGS images into a two-dimensional space using t-SNE~\cite{van_der_maaten_visualizing_2008}, separately for the base model, the meta-learned model, and a human similarity embedding derived from the THINGS odd-one-out dataset~\cite{hebart_revealing_2020, hebart_things-data_2023} (\autoref{fig:geoms}A-C). The base and meta-learned embeddings are clearly different: the most notable change after meta-learning is a sharper separation of organic objects (e.g., animals, plants, and food) from inorganic ones (e.g., clothing, furniture, and vehicles), an exaggerated separation that is also present in the human similarity embedding.

To quantitatively confirm this observation, we compared the representations of our models against the 66-dimensional human similarity embedding using Centered Kernel Alignment (CKA)~\cite{kornblith_similarity_2019}. Across all tested backbones, meta-learned representations showed a substantial increase in CKA with the human similarity space compared to their respective base models (\autoref{fig:geoms}D). Further analysis of class separation confirmed that meta-learning significantly improved the distinctness of both coarse-grained semantic domains (\autoref{fig:geoms}E) and fine-grained object categories (\autoref{fig:geoms}F). These results demonstrate that meta-learning on sparse, high-level tasks not only improves behavioural and neural alignment, but fundamentally reshapes the representational geometry to mirror the categorical and hierarchical organisation of human visual cognition.

\section*{Discussion}
\label{sec:discussion}
Why are human visual representations organised the way they are? Our results suggest that the functional demand to remain adaptable across a broad space of semantically rich tasks is itself a powerful organising force. The mechanism is not learning any particular set of tasks but meta-learning, being required to learn novel tasks from limited data on-the-fly. This pressure for flexible reuse reconfigures visual representations to mirror human similarity judgements, learning behaviour, and the functional organisation of the high-level visual cortex, all without any direct supervision on human data. Notably, this reorganisation was not reproduced by a multitask model trained on exactly the same distribution of tasks, pointing to learning-to-learn itself, rather than mere task exposure, as the active ingredient.

Understanding the factors that shape human vision has broadly been pursued along three axes in neural network modelling: architecture, data, and the computational problem being solved. On the architectural side, brain-like representations have been linked to hierarchical processing constraints~\cite{kubilius_brain-like_2019} and recurrence~\cite{kietzmann_recurrence_2019}. On the data side, developmental input statistics support the grounded emergence of concepts~\cite{vong_grounded_2024}, and matching image distributions to human experience improves alignment~\cite{mehrer_ecologically_2021}. Our work addresses the third axis. Prior accounts have shown that human-like structure can emerge when networks are optimized for particular broad functions, such as visual compression~\cite{konkle_self-supervised_2022}, self-supervised prediction~\cite{bakhtiari_functional_2021}, or modelling visual dynamics~\cite{tang_diverse_2025}. These objectives demonstrate that representational geometry is shaped by functional demands. Our proposal differs in the nature of that demand: rather than optimizing representations for a fixed target function over images, meta-learning optimizes them for the rapid acquisition of new target functions from limited context. In this sense, human-like geometry emerges not because the model is trained to solve any one semantic problem, but because it is trained to make many such problems quickly learnable.

Not just any form of meta-learning produces this effect on behaviour. Our comparisons showed that both the disentangled structure and the high level of the task distribution are critical for behavioural alignment: training on entangled activations at the same level of abstraction, or on sparse features at a lower level, did not yield comparable improvements. Brain alignment showed a more heterogeneous picture across these ablations, suggesting the task distribution shapes the semantic format of the representations more than their gross fit to cortex. What matters is not just learning to learn, but learning to learn over the right kind of conceptual vocabulary, one that is interpretable and semantically rich.

This specificity is mirrored in the neural data. Meta-learning selectively improved alignment with high-level visual cortex, category-selective regions for bodies and faces, as well as object- and scene-selective areas, while leaving early visual areas unchanged. This pattern is consistent with the idea that the pressure to flexibly acquire semantically rich tasks reshapes the representations that support categorical and conceptual processing, the regions most engaged when humans navigate the kind of abstract, context-dependent distinctions.

Previous work has established meta-learning as a powerful framework for modelling cognitive functions and phenomena~\cite{binz_meta-learned_2024}, including cognitive biases~\cite{dasgupta_theory_2020}, decision-making~\cite{binz_heuristics_2022}, and language understanding~\cite{mccoy_universal_2020}. These studies use meta-learning to capture how people behave. Our contribution is different: we show that the pressures meta-learning places on representations are sufficient to reorganise their geometry into one that mirrors human visual cognition. The goal is not solely to maximise alignment with human data, models directly fine-tuned on behavioural benchmarks will typically score higher on those benchmarks, but to offer a normative account of why human-like organisation emerges in the first place, from general functional demands alone.

One limitation is that our task generation depends on existing neural networks. SAEs decompose activations into interpretable units, but the resulting vocabulary inherits the biases of the encoders used to produce them; the semantically rich tasks we meta-learn over are therefore tied to the kinds of concepts CLIP is equipped to represent. A second limitation is that our training and evaluation rely on object-centric naturalistic image sets; we do not test whether the same principles hold for scene-level, dynamic, or non-visual input. Whether similar principles hold when the conceptual vocabulary is grounded in non-engineered sources, for instance through interaction with physical environments or linguistic input, remains an open question. More broadly, our framework generates testable predictions: if the demand to learn diverse semantically rich tasks is what shapes representations, then differences in the richness or structure of the tasks an agent encounters should produce measurable differences in representational geometry. Varying properties of the task distribution, such as task diversity, conceptual density, and environmental noise, will help map out the normative conditions that act on human representations.

Our findings suggest that the rich, flexible nature of human visual representations is not a by-product of architectural scale or training data, but a predictable consequence of the pressure to remain adaptable.

{\small
\section*{Methods}
\label{sec:methods}
\subsection*{Task distributions}

Training a meta-learning model requires a distribution of tasks. To generate semantically meaningful tasks over images at scale, we employed pretrained SAEs~\cite{cunningham_sparse_2023} available through \texttt{ViT-Prisma}~\cite{joseph_vit_2023, joseph_prisma_2025}. For our main experiments, we used a Top-K SAE~\cite{gao_scaling_2025} trained on the \texttt{[CLS]} token representations at layer $11$'s residual stream of OpenCLIP ViT-B-32~\cite{ilharco_openclip_2021} on ImageNet~\cite{deng_imagenet_2009}, encoding each image with $64$ active latents. Each SAE latent defines one task: an image is a positive example if that latent is active for it, with the activation magnitude serving as the graded outcome $y_t$. We extracted these representations for COCO~\cite{lin_microsoft_2014} images, using the train split for training and the validation and test splits for evaluation. Crucially, none of these images were later used in human-alignment evaluations. For all feature types, we filtered out latents with fewer than $120$ non-zero activations across all images, as these provide insufficient positive examples for learning.

To further test the role of interpretability and abstraction level, we created two additional task distributions. To test the role of the SAE's sparse, interpretable structure, we used the raw residual stream representations of the CLIP model at layer $11$ directly, keeping the level of abstraction constant while removing that structure. To isolate abstraction level, we trained a separate SAE on the residual stream of layer $6$ of the same model, keeping the sparse structure while shifting to lower-level features. Summary statistics for all three distributions are given in Table \ref{tab:tasks}.

\begin{table}[h]
  \centering
  \begin{tabular}{lrr}
  \toprule
  Features & Train tasks & Eval tasks \\
  \midrule
  SAE (Layer 11)     & 7,251 & 2,947 \\
  SAE (Layer 6)      & 499   & 459   \\
  ViT Residual (Layer 11) & 768   & 768   \\
  \bottomrule
  \end{tabular}
  \caption{Number of training and evaluation tasks per feature type. The underlying pool of $118{,}287$ training images and $45{,}670$ evaluation images is shared across all three conditions.}
  \label{tab:tasks}
  \end{table}

\subsection*{Model architecture \& training}

In each episode, a task was sampled uniformly at random from the available set. Images were then drawn without replacement to form a sequence of length $T$: the number of positive examples $n_+$ was drawn from $\mathcal{N}(T/2,\ T/20)$ and clamped to the number of available positives, with the remainder as negatives; the full sequence was randomly shuffled. For SAE features, an image was positive if its activation for the sampled latent was non-zero, and the graded outcome $y_t$ was the raw activation value; for the raw residual stream, an image was positive if its activation on the sampled dimension exceeded the across-image median, and $y_t$ was the signed deviation from that median. The meta-learner received a sequence of inputs $o_{1:T}$ and was trained to predict a corresponding sequence of outputs $y_{1:T}$. At each trial $t$, the model observed $o_t = [x_t; y_{t-1}]$, where $x_t = \phi(I_t)$ was the encoder representation of image $I_t$ and $y_{t-1}$ was the outcome of the previous trial (zero at $t=1$). This was projected to the original representation dimensionality via a learned linear embedding (which improves the efficiency of in-context learning; see Supplementary Section 2 and Fig. S2) and processed by a causally-masked attention-only Transformer~\cite{vaswani_attention_2017} with $1$ layer, $16$ attention heads, Rotary Positional Embeddings~\cite{su_roformer_2024}, and pre-attention layer normalisation~\cite{ba_layer_2016}. For evaluations demonstrating that performance remains robust across alternative sequence architectures (including LSTMs~\cite{hochreiter_long_1997} and deeper Transformers), see Supplementary Fig. S1. The output at each position was passed to two linear heads: one predicting whether the current image was a positive instance of the sampled task (binary cross-entropy loss), and one predicting the graded outcome (mean-squared-error loss; for SAE features computed on positive examples only, since negatives have zero magnitude by definition; for raw residual features computed on all examples). The two losses were combined using uncertainty weighting~\cite{kendall_multi-task_2018}, which learned a precision per task to automatically balance their contributions.

Models were trained for up to $10{,}000$ steps with a batch size of $4{,}096$ and sequence length $T=120$. We used the schedule-free AdamW optimizer~\cite{defazio_road_2024} with learning rate $2.5 \times 10^{-4}$ and no weight decay. Training used mixed-precision (bfloat16) and was distributed across $4$ GPUs. We evaluated every $100$ steps on a held-out set of episodes and applied early stopping with a patience of $1{,}000$ steps, based on the mean of classification accuracy and $R^2$ on activation magnitudes. Early stopping was only applied once the model exceeded $60\%$ classification accuracy, to avoid premature termination before the model had meaningfully begun to learn.

\paragraph{Pretrained image encoders} We evaluated four frozen pretrained encoders $\phi$: CLIP ViT-B-32~\cite{ilharco_openclip_2021}, DINOv3 ViT-B/16~\cite{simeoni_dinov3_2025}, SigLIP2 ViT-B/16~\cite{tschannen_siglip_2025}, and a ViT-L trained with masked image modelling (MAE) at scale~\cite{fan_scaling_2025}. These span contrastive (CLIP, SigLIP2), self-supervised distillation (DINOv3), and masked autoencoding (MAE) training objectives. CLIP was additionally included as it underlies the SAE used to define the task distributions. For CLIP, $x_t$ is the layer $11$ CLS-token residual stream ($768$-d), the same space in which the SAE latents are defined; for DINOv3 and SigLIP2, we used the model-default pooled representation from \texttt{timm}~\cite{wightman_pytorch_2019} ($768$-d); for the MAE, we used the mean of patch token representations ($1{,}024$-d).

\paragraph{Multitask learning} To isolate the contribution of the learning-to-learn pressure from mere exposure to the task distribution, we trained a non-sequential multitask model that receives the same training signal as the meta-learner but without episodic structure. Each encoder representation $x_t = \phi(I_t)$ was passed through a learned linear layer of the same dimensionality as the meta-learner's input projection, and the resulting shared representation was fed to a set of per-task linear heads -- one per task in the training set. Each head produced the same two predictions as the meta-learner (positive-instance classification and graded outcome), trained with the same binary cross-entropy and mean-squared-error losses combined via uncertainty weighting~\cite{kendall_multi-task_2018}. At each training step, we sampled (image, task) pairs; the model never receives feedback from previous trials and never has to infer which task is active, since each task has its own dedicated head. We used the same optimiser, learning rate, batch size, and early-stopping criterion as for the meta-learner. For downstream evaluations, we extracted model representations by applying the shared linear layer to encoder features, mirroring the procedure used to extract meta-learned representations from the meta-learner.

\subsection*{Behavioural evaluations}

For all behavioural evaluations, we extracted a static meta-learned representation for each image by applying the meta-learner's learned input projection to the encoder features, independently per image with no sequence context. Performance was measured as the negative-log-likelihood (NLL) of human choices under a per-participant softmax temperature model. For interpretability, we converted the NLLs into McFadden's $R^2$~\cite{mcfadden_conditional_1972} using the following formula:

$$
\text{McFadden's} \: R^2 = 1 - \dfrac{NLL_{Model}}{NLL_{Random}}
$$

where a value of $1$ corresponds to a theoretically perfect model and a value of $0$ to one at chance-level.

\paragraph{Odd-one-out (THINGS)} We evaluated human alignment using the THINGS triplet dataset~\cite{hebart_things-data_2023}, comprising $4{,}699{,}160$ triplets collected from $12{,}340$ participants across the $1{,}854$ object categories in THINGS~\cite{hebart_things_2019}. On each trial, participants chose the odd one out among three images. We computed pairwise cosine similarities between image representations within each triplet. The model's predicted odd-one-out is the image whose removal leaves the most similar pair (i.e., the logit for each image equals the cosine similarity between the remaining two) and fit a softmax model with per-participant temperature via 5-fold cross-validation.

\paragraph{Odd-one-out (Levels)} We additionally evaluated on the Levels odd-one-out dataset~\cite{muttenthaler_levels_2024, muttenthaler_aligning_2025}, comprising $123{,}043$ trials from $448$ participants on ImageNet images. Triplets varied in composition: within-class (all three images from the same class), class-border (two from one class, one from another), and between-class (all three from different classes), enabling evaluation across varying levels of visual similarity. Evaluation followed the same procedure as above.

\paragraph{Category learning} We evaluated the models against human performance on a sequential category learning task~\cite{demircan_evaluating_2024} ($N = 91$, $120$ trials, $3$ semantic dimensions). Each participant was randomly assigned one of three latent semantic dimensions (e.g.\ whether an object is metallic); images were drawn from THINGS and labelled accordingly. Participants saw images one at a time, predicted which of two categories each image belonged to, and received feedback, unbeknown to them that the decision rule was semantic. We simulated this process by fitting an online L2-regularised logistic regression to the meta-learned representations: at each trial $t$, the model is fit on all preceding trials $1{:}t{-}1$ and used to predict trial $t$. The regularisation strength $C$ was selected via 5-fold cross-validation on the full sequence. Predictions are uniform before both classes have been observed.

\paragraph{Reward learning} We evaluated the models against human performance on a sequential reward learning task~\cite{demircan_evaluating_2024, demircan_decision-making_2022} ($N = 82$, $60$ trials, $3$ semantic dimensions). Each participant was randomly assigned one of three latent semantic dimensions; reward magnitudes were defined over that dimension. On each trial, participants chose which of two simultaneously presented THINGS images had higher reward; all rewards were then revealed. We simulated this by fitting an online Bayesian Ridge regression: at each trial $t$, the model trained on trials $1{:}t{-}1$ predicts reward for each option, and the difference in predicted values serves as the choice logit; the model is then updated to include trial $t$.

\subsection*{Brain evaluations}

We evaluated brain alignment using the THINGS-fMRI dataset~\cite{hebart_things-data_2023} ($N = 3$), in which participants viewed images from the THINGS object concept database across $12$ sessions. We retained only voxels with a single-trial noise ceiling above $5\%$ and report noise-ceiling corrected $R^2$. We fit a Ridge regression model using nested cross-validation: an outer leave-one-session-out loop for evaluation, with $\alpha$ selected independently per voxel via inner cross-validation ($\alpha \in [10^{-6}, 10^3]$) on the training fold. Encoder (base) representations served as a baseline. We report results across $11$ ROIs spanning early visual cortex (V1, V2, V3, hV4), object- and scene-selective regions (LOC, PPA, RSC, TOS), and category-selective regions (EBA, FFA, OFA).

\subsection*{Statistical comparisons}

\paragraph{Behavioural evaluations} For all behavioural evaluations, we compared models using Group Bayesian Model Comparison~\cite{rigoux_bayesian_2014}, treating each participant's total NLL as the model log-likelihood. This estimates the frequency with which each model best explains the data across the group and reports the protected exceedance probability (PXP), the probability that a given model is the most likely.

\paragraph{Brain evaluations}  At the ROI level, we used an exhaustive sign-flip test across the $12$ leave-one-session-out folds ($2^{12}$ permutations) per participant, combined $p$-values across participants using Fisher's method, and corrected for multiple comparisons across ROIs using the Benjamini-Hochberg procedure.

\subsection*{Geometric analyses}

To characterise how meta-learning transforms the representational geometry, we compared base and meta-Learned representations on THINGS images along two dimensions.

\paragraph{Alignment with human similarity geometry} We measured alignment with the 66-dimensional SPoSE human similarity embedding~\cite{hebart_revealing_2020, hebart_things-data_2023} using Centered Kernel Alignment (CKA)~\cite{kornblith_similarity_2019}, computed on the first image per THINGS category ($N = 1{,}854$), matched to the SPoSE ordering. Estimates and standard errors were obtained via bootstrap resampling of observations ($n = 100$).

 \paragraph{Class separation} We measured class separation as $1 - \sigma^2_\text{within} / \sigma^2_\text{total}$, where $\sigma^2_\text{within} = \sum_c \sum_{i \in c} |x_i - \mu_c|^2$ is the total within-class variance and $\sigma^2_\text{total} = \sum_i |x_i - \mu|^2$ is the total variance, computed on L2-normalised representations~\cite{kornblith_why_2021}. For fine-grained separation, we used all THINGS images with the $1{,}854$ object categories as labels. For coarse-grained separation, we used one image per category ($N = 1{,}854$) with $26$ binary semantic category labels from THINGS metadata. Estimates and standard errors were obtained via bootstrap resampling ($n = 100$).

\section*{Acknowledgement}
This work was supported by Helmholtz Munich and the European Research Council (ES).

\section*{Code and data availability}
\label{sec:dataave}
All code and data needed to reproduce the results reported in this manuscript are publicly available at \url{https://github.com/candemircan/metalign} and \url{https://osf.io/dren7} respectively.

\section*{Competing Interests Statement}
The authors declare no competing interests.

\bibliography{references.bib}
}

\clearpage

\setcounter{section}{0}
\setcounter{figure}{0}
\setcounter{table}{0}
\renewcommand{\thefigure}{S\arabic{figure}}
\renewcommand{\theequation}{S\arabic{equation}}
\renewcommand{\thetable}{S\arabic{table}}

\begin{center}
{\Large\bfseries Supplementary Information}\\[0.5em]
{\large Meta-learning as a principle for human-like visual representations}
\end{center}
\vspace{0.5cm}

\section{Robustness to architectural variations}

We investigated whether the emergence of human-like representations was sensitive to the specific choice of the sequence model. We tested several variations, including varying the number of attention heads, changing the depth and the expressivity of the model, changing optimizer settings, as well as replacing the transformer with an LSTM~\cite{hochreiter_long_1997}. Overall, there were very few changes in alignment as a result of these modifications in most cases. The results are shown in Supplementary \autoref{fig:architectures}.

\section{Is linear projection useful for meta-learning?}

A key component of our architecture is the learned linear projection of encoder features before they enter the causal Transformer. We evaluated the model's ability to learn in-context with and without this projection. Models with the linear embedding achieved higher asymptotic accuracy across context positions. The Mean Squared Error (MSE) was significantly lower for the same models. This suggests that a simple linear reconfiguration of the feature space is a highly effective way to prepare representations for rapid, in-context semantic learning. The results are shown in Supplementary \autoref{fig:linear}.

\section{Multitask model}

To isolate the contribution of the learning-to-learn pressure from mere exposure to the task distribution, we trained a non-sequential multitask model alongside the meta-learner. The model receives exactly the same image-task supervision as the meta-learner but processes each image independently through a shared linear projection followed by per-task linear heads, and is therefore never required to infer the active task within an episode. Full architectural and optimisation details are given in the main Methods (Section~\textit{Multitask model}).

Across all four encoders, meta-learning yielded significantly higher noise-ceiling corrected $R^2$ than multitask learning in every high-level ROI (EBA, FFA, OFA, PPA, RSC, TOS, LOC; all $p < .001$, exhaustive sign-flip tests combined across participants via Fisher's method, Benjamini--Hochberg corrected across ROIs; Supplementary \autoref{fig:meta_supp}). This mirrors the behavioural comparison in Fig.~3 of the main text and reinforces the interpretation that the learning-to-learn pressure itself, rather than mere exposure to the task distribution, is what reorganises representations in a way that better matches high-level visual cortex.

\section{Brain alignment across image encoders}

To confirm the generality of our findings in the brain, we extended our fMRI predictivity analysis to all four image encoders: SigLIP2, MAE, DINOv3, and CLIP. Meta-learning consistently increased $R^2$ in category-selective areas (EBA, FFA, OFA) and scene-selective areas (PPA, RSC, TOS) across all backbones. These results are shown in Supplementary \autoref{fig:brain_combined}.

\section{Brain alignment across task distributions}

We compared brain alignment when training on different task distributions (SAE Layer 11, Raw, and SAE Layer 6). Here, we observed more heterogeneous effects of the task distribution, which are displayed in Supplementary \autoref{fig:brain_ablations}.

\begin{figure}[h!]
    \centering
    \includegraphics[width=.9\textwidth]{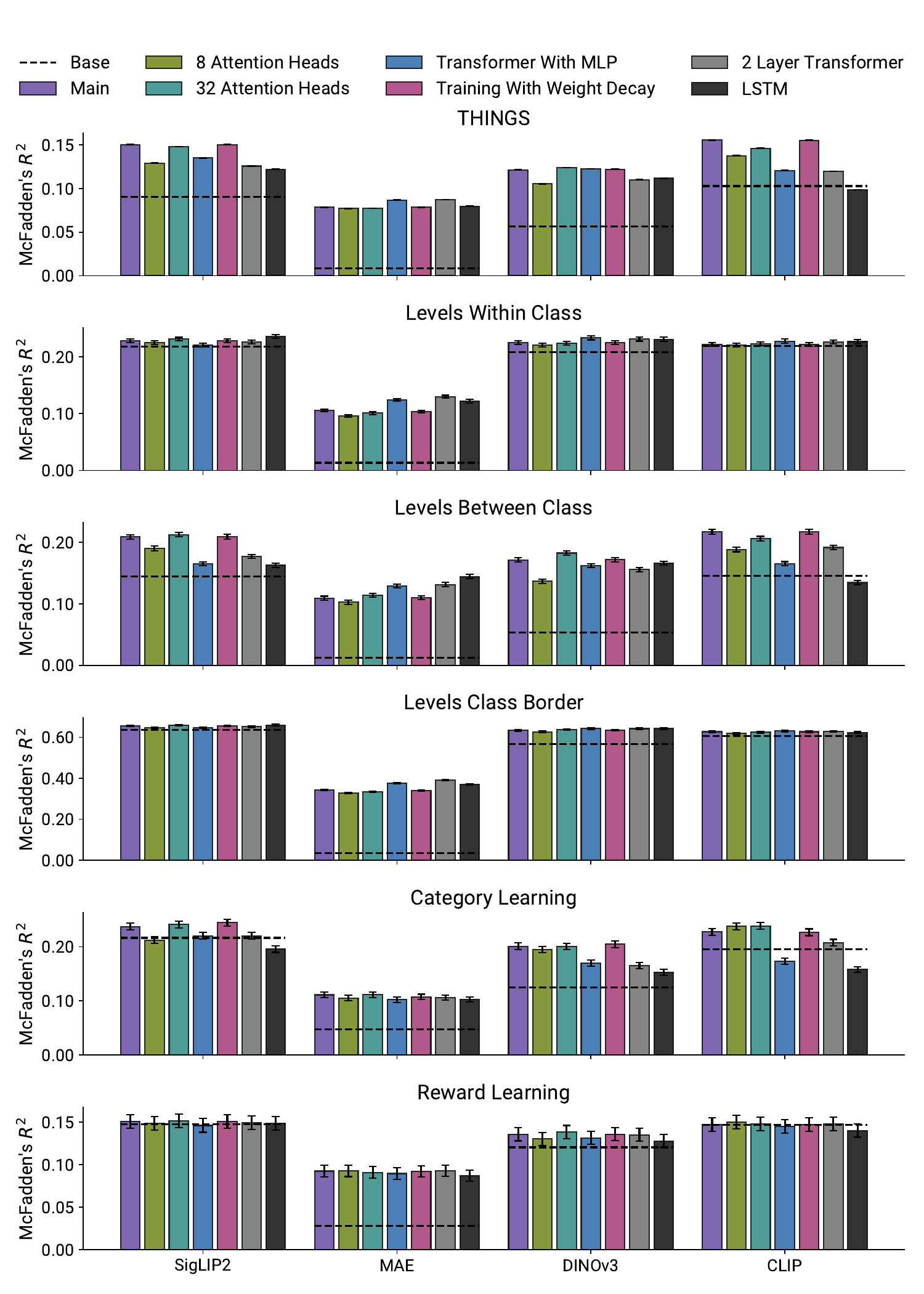}
    \vspace{-1.2cm}
    \singlespacing  \caption{ \footnotesize
    \textbf{Robustness to architectural variations}. Alignment performance (McFadden’s $R^2$) across behavioural tasks for various sequence model architectures, including LSTMs and Transformer variants with different depths, attention heads, and MLP blocks. The dashed line represents the performance of the base model. Results demonstrate that meta-learning gains are mostly robust to the specific choice of the sequence model.
    }
    \label{fig:architectures}
\end{figure}

\begin{figure}[h!]
    \centering
    \includegraphics[width=\textwidth]{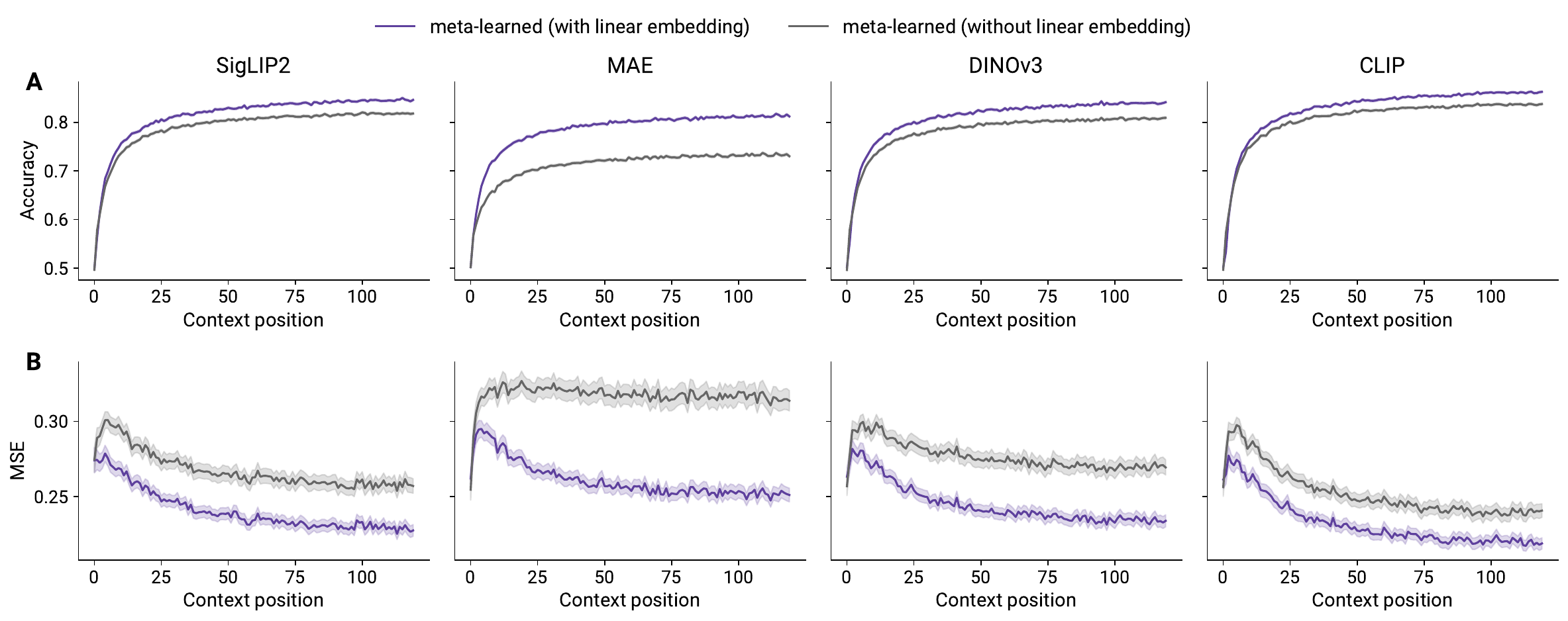}
    \vspace{-1.2cm}
    \singlespacing  \caption{ \footnotesize
    \textbf{Impact of linear projection on in-context learning.} (\textbf{A}) Accuracy and (\textbf{B}) Mean Squared Error (MSE) as a function of context position. Blue lines indicate meta-learned models with linear embedding; gray lines indicate models without it. The linear projection facilitates faster and more accurate in-context learning across all tested backbones.
    }
    \label{fig:linear}
\end{figure}

\begin{figure}[h!]
    \centering
    \includegraphics[width=\textwidth]{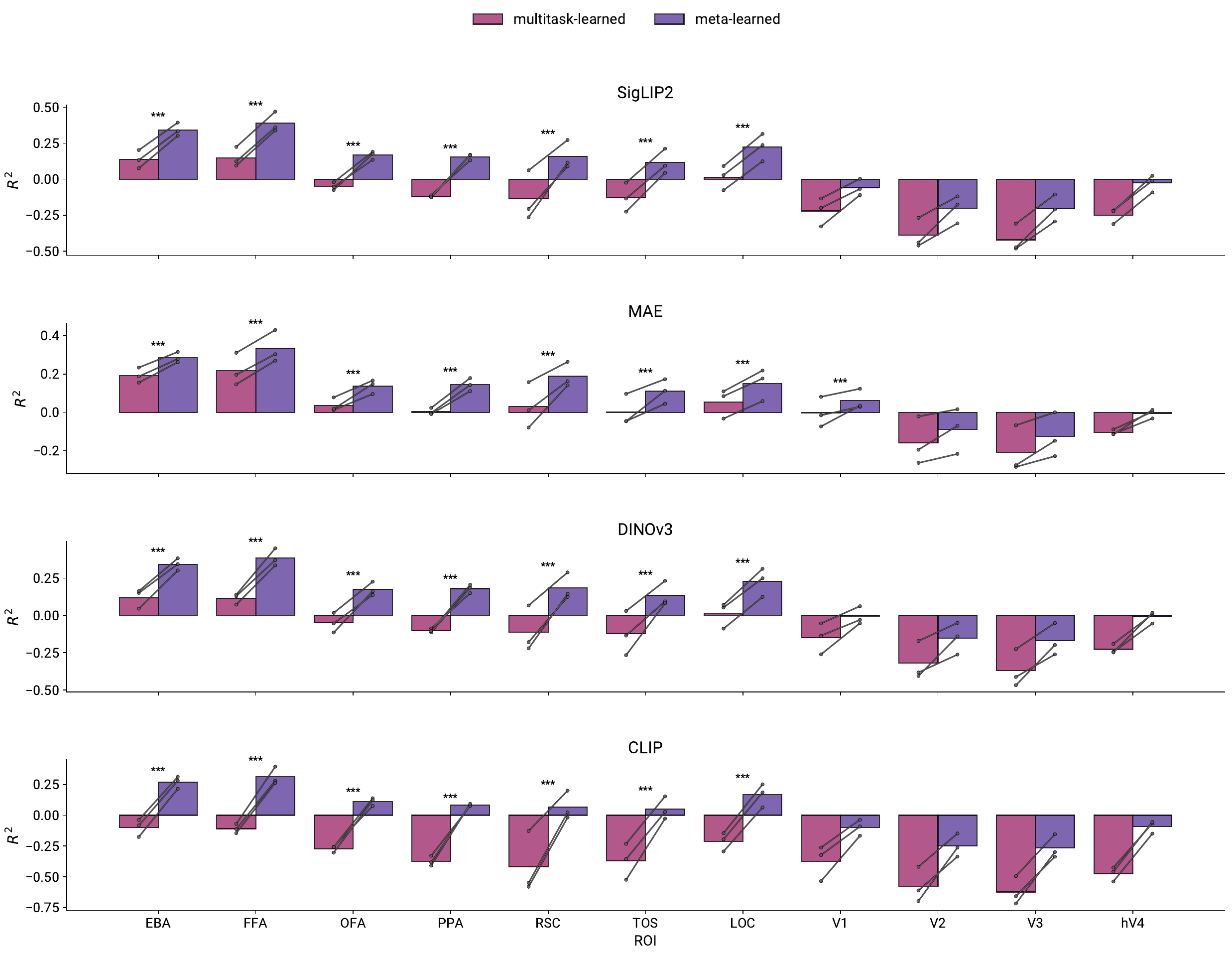}
    \vspace{-1.2cm}
    \singlespacing  \caption{ \footnotesize
    \textbf{Meta-learning outperforms multitask learning in predicting high-level visual cortex.} Noise-ceiling corrected $R^2$ for multitask-learned (pink) and meta-learned (purple) versions of SigLIP2, MAE, DINOv3, and CLIP across 11 regions of interest (ROIs). Statistical significance ($***$) indicates a significant improvement in predictive power for meta-learning over multitask learning. Meta-learning consistently outperforms multitask learning across all high-level ROIs for all four encoders. Individual data points represent the average $R^2$ per participant (N = 3).
    }
    \label{fig:meta_supp}
\end{figure}

\begin{figure}[h!]
    \centering
    \includegraphics[width=.7\textwidth]{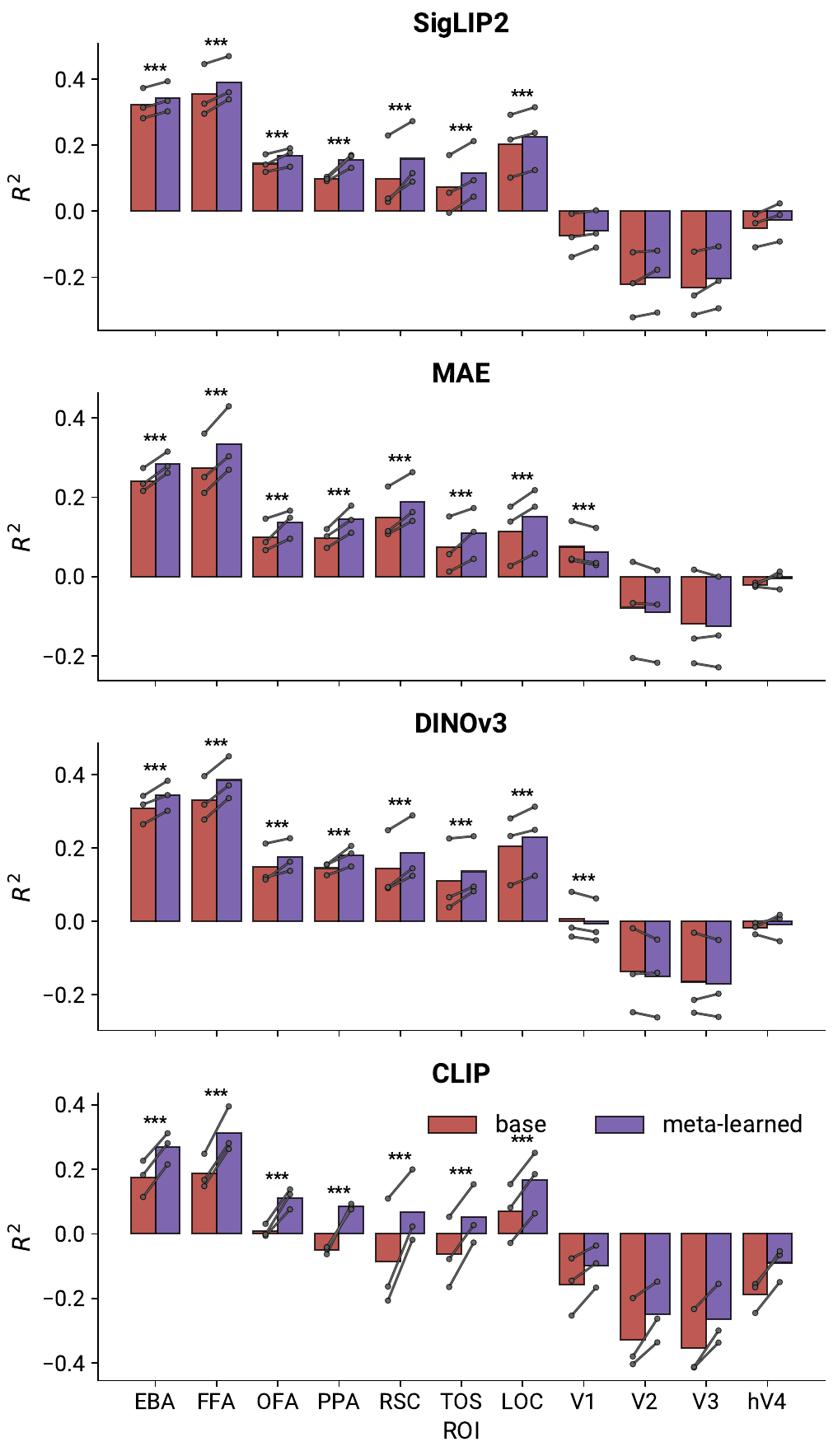}
    \vspace{-1.2cm}
    \singlespacing  \caption{ \footnotesize
    \textbf{Brain alignment across all backbones.} Noise-ceiling corrected $R^2$ for base (red) and meta-learned (purple) versions of SigLIP2, MAE, DINOv3, and CLIP across 11 regions of interest (ROIs). Statistical significance ($***$) indicates a significant improvement in predictive power following meta-learning, particularly in high-level regions selective for bodies, faces, objects, and scenes. Individual data points represent the average $R^2$ per participant (N = 3).
    }
    \label{fig:brain_combined}
\end{figure}

\begin{figure}[h!]
    \centering
    \includegraphics[width=.7\textwidth]{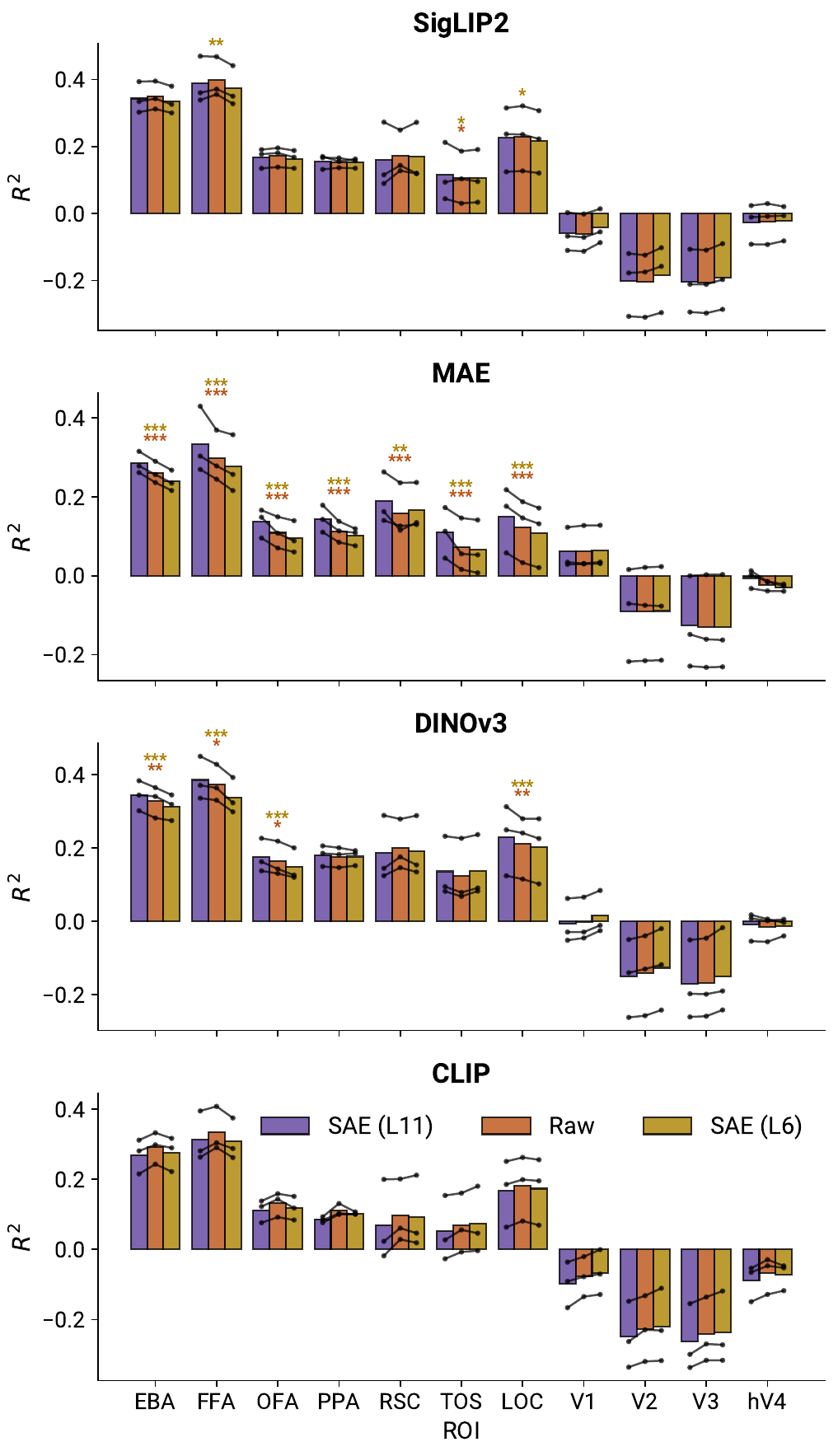}
    \vspace{-1.2cm}
    \singlespacing  \caption{ \footnotesize
    \textbf{Brain alignment across task distributions.} Comparison of predictive power ($R^2$) when models are meta-trained on different task distributions: high-level disentangled (SAE L11), high-level entangled (Raw), and mid-level (SAE L6). Results paint a mixed picture for how the properties of the task distribution affect brain alignment, suggesting that the brain-alignment benefit of meta-learning is driven primarily by the episodic training structure rather than by properties of the task distribution.
    }
    \label{fig:brain_ablations}
\end{figure}

\end{document}